\def\year{2022}\relax
\documentclass[letterpaper]{article} 
\usepackage{aaai22}  
\usepackage{times}  
\usepackage{helvet}  
\usepackage{courier}  
\usepackage[hyphens]{url}  
\usepackage{graphicx} 
\usepackage{natbib}  
\usepackage{caption} 
\DeclareCaptionStyle{ruled}{labelfont=normalfont,labelsep=colon,strut=off} 
\usepackage{xspace}
\newcommand*{\eg}{e.g.\@\xspace}
\newcommand*{\ie}{i.e.\@\xspace}
\usepackage{subfigure}
\usepackage{bm}
\usepackage{amsmath}
\usepackage{amssymb}
\usepackage{enumitem}
\usepackage{booktabs}
\usepackage{multirow}
\usepackage{setspace}
\usepackage[ruled,vlined,linesnumbered]{algorithm2e}
\newcounter{algoline}
\newcommand\Numberline{\refstepcounter{algoline}\nlset{\thealgoline}}
\usepackage{makecell}
\usepackage{algpseudocode}
\usepackage[table]{xcolor}
\usepackage{pdfpages}
\usepackage{hyperref}
\hypersetup{
    colorlinks = true,
    linkcolor=black,
    filecolor=blue,      
    urlcolor=blue,
    citecolor=black,
}
\definecolor{llgray}{rgb}{0.9, 0.9, 0.9}

\usepackage{newfloat}
\usepackage{listings}
\begin{document}


\title{Deconfounded Visual Grounding}

\author{
Jianqiang Huang\textsuperscript{\rm 12},
Yu Qin\textsuperscript{\rm 2},
Jiaxin Qi\textsuperscript{\rm 1},
Qianru Sun\textsuperscript{\rm 3},
Hanwang Zhang\textsuperscript{\rm 1}
}
\affiliations{
\textsuperscript{\rm 1}Nanyang Technological University, Singapore 
\textsuperscript{\rm 2}Damo Academy, Alibaba Group
\textsuperscript{\rm 3}Singapore Management University\\
jianqiang.jqh@gmail.com, dongdong.qy@alibaba-inc.com, jiaxin003@e.ntu.edu.sg\\ qianrusun@smu.edu.sg, hanwangzhang@ntu.edu.sg
}

\maketitle

\begin{abstract}
We focus on the confounding bias between language and location in the visual grounding pipeline, where we find that the bias is the major visual reasoning bottleneck. For example, the grounding process is usually a trivial language-location association without visual reasoning, \eg, grounding any language query containing \texttt{sheep} to the nearly central regions, due to that most queries about \texttt{sheep} have ground-truth locations at the image center. First, we frame the visual grounding pipeline into a causal graph, which shows the causalities among image, query, target location and underlying confounder. Through the causal graph, we know how to break the grounding bottleneck: deconfounded visual grounding. Second, to tackle the challenge that the confounder is unobserved in general, we propose a confounder-agnostic approach called: Referring Expression Deconfounder (RED), to remove the confounding bias. Third, we implement RED as a simple language attention, which can be applied in any grounding method. On popular benchmarks, RED improves various state-of-the-art grounding methods by a significant margin. Code will soon be available at: \url{https://github.com/JianqiangH/Deconfounded_VG}.
\end{abstract}

\section{Introduction}
\label{sec:1}
Visual Grounding, also known as the task of Referring Expression Comprehension~\cite{karpathy2014deep,karpathy2015deep,ref-split}, has greatly expanded the application domain of visual detection, from a fixed noun vocabulary, \eg, \texttt{dog}, \texttt{car}, and \texttt{person}, to free-form natural language on demand, \eg, \emph{the dog next to the person who is driving a car}. Regarding the place of candidate proposal generation in the grounding pipeline, there are two traditional camps: two-stage and one-stage. For the two-stage, the 
first detection stage denotes any object detectors~\cite{ren2015faster} for extracting candidate regions from the image, and the second visual reasoning stage is to rank the candidates and select the top one based on their similarities to the query embedding~\cite{yu2018mattnet, liu2019learning}; for the one-stage, the detection and reasoning is unified by directly performing the referent detection: generating the referent region proposals with confidence scores and spatial coordinates for each pixel in the multi-modal feature map, which is usually the concatenation of the visual feature map, location coordinates, and language embeddings~\cite{yang2019fast, yang2020improving}. However, if the gradients can backpropagate through the visual detector in the two-stage methods, there is little difference between the two camps.
Without loss of generality, this paper will be focused on one-stage methods. In particular, the introduction of our approach will be based on the one-stage model~\cite{yang2019fast} with YOLO~\cite{yolov4} backbone which has been validated of high inference speed and good generalizability.
\begin{figure}[t]
\centering
\includegraphics[width=3.2in]{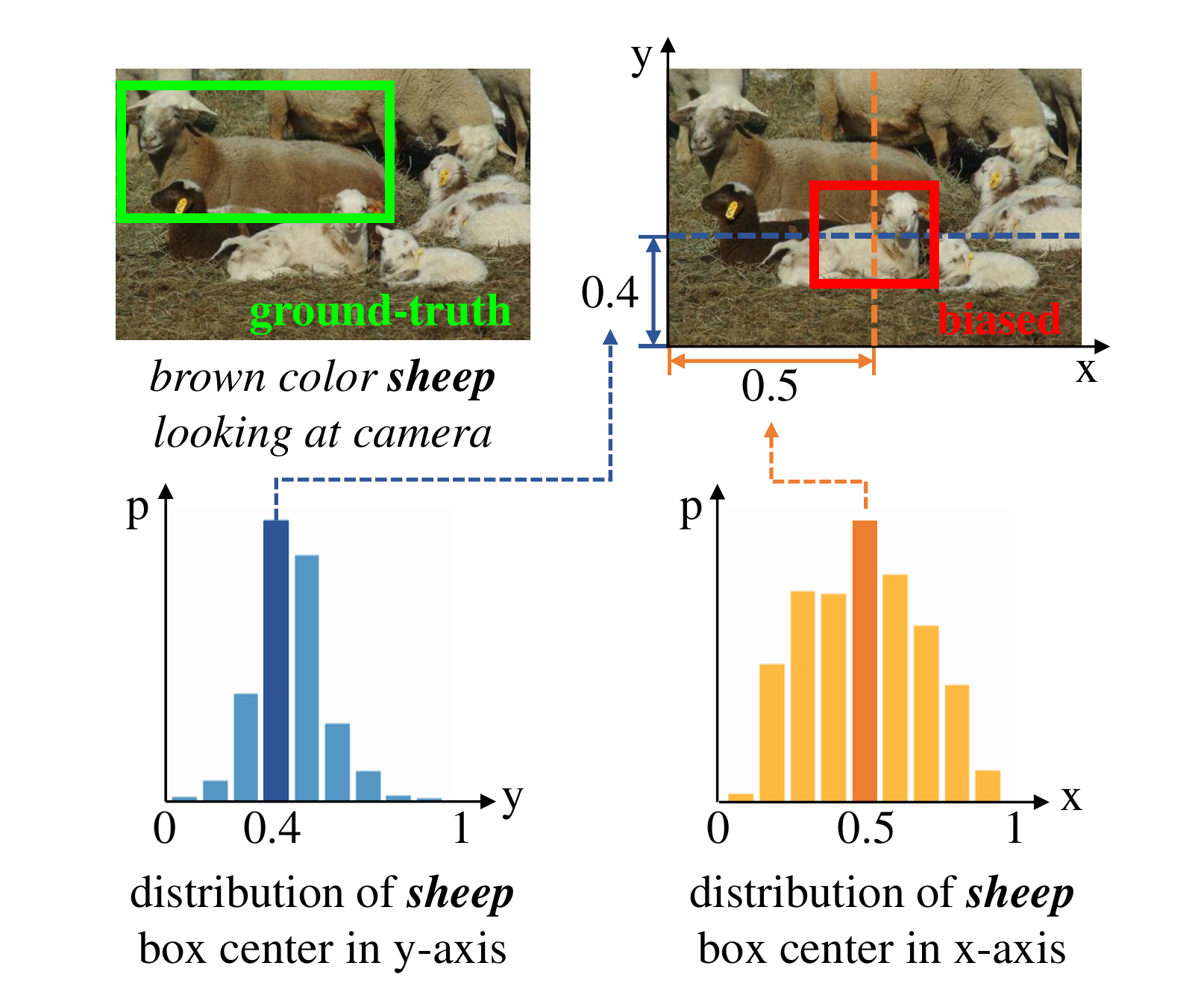}
\caption{Two image examples: one with ground truth location of \emph{brown color sheep looking at camera}; and the other with a wrong prediction caused by the language bias in the grounding model.
Blue bars denote the distribution of the center positions of \emph{sheep} on the y-axis (of the image), and orange bars for the x-axis (of the image).}
\label{fig:teaser}
\end{figure}

\begin{figure}[t]
\centering
\includegraphics[width=3.2in]{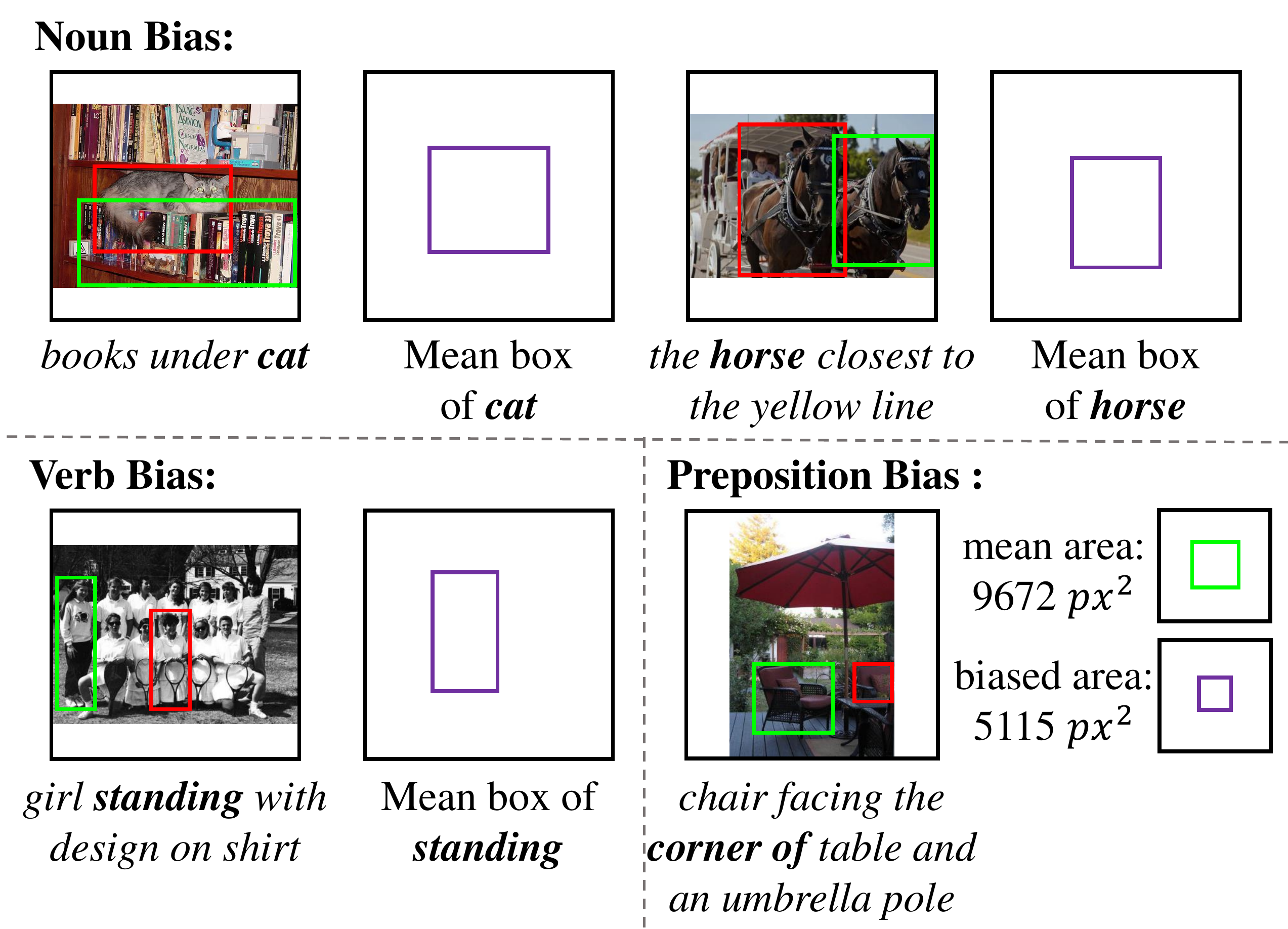}
\caption{
Three types of language bias observed in the grounding model~\cite{yang2019fast}. The noun and verb biases are from the model trained on RefCOCO+~\cite{refcoco} and the preposition bias is from RefCOCOg~\cite{refcocog}.
Green boxes denote ground-truth, red boxes denote biased prediction and purple boxes denote language biased location. ``Mean box'' denotes the average region over all ground truth bounding boxes of a specific language fragment (\eg, \texttt{horse}) in the dataset.
For \texttt{corner of}, ``mean area''
is the average size of all ground truth boxes
and "Biased" one is the ground truth boxes averaged over samples with a specific language fragment (\eg, ``corner of'').}
\label{fig:1}
\end{figure}
\begin{figure*}[htbp]
\centering
\includegraphics[width=6.3in]{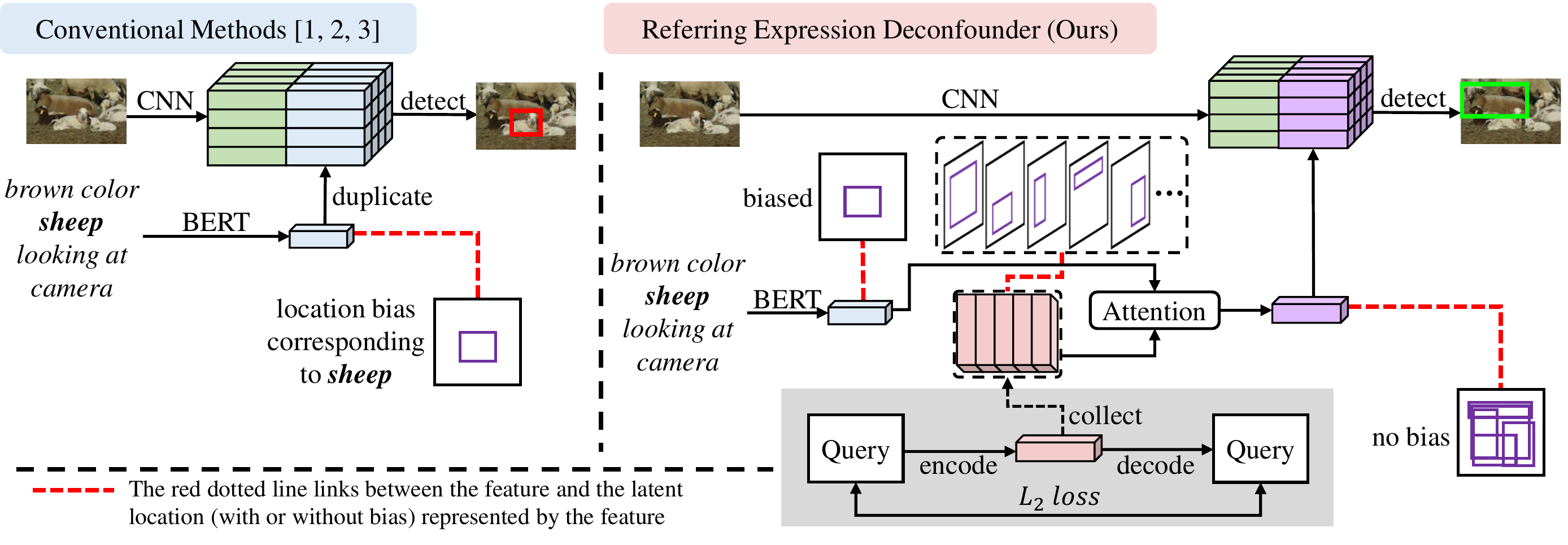}
\caption{The pipeline of conventional grounding methods vs. our RED. [1, 2, 3] denote the methods in \cite{yang2019fast}, \cite{yang2020improving} and \cite{lbyl}, respectively. In the left part, green bars denote different region features and blue bars denote the duplicated language features.
In the right part, the gray region denotes the process of dictionary extraction before training any grounding model. ``Query'' denotes all the referring expressions in the current training grounding dataset.
The dashed black box bounding multiple pink bars represents the dictionary we get by clustering all ``pink features'' extracted via a trained auto-encoder (in the gray region). The purple bars denote the deconfounded language features derived by our RED.
}
\label{fig:2}
\end{figure*}

By analyzing the failure examples of existing grounding methods, we find an ever-overlooked bias, as illustrated in Figure~\ref{fig:teaser}: 
the bounding box prediction tends to be strongly biased to some particular position, \eg, for the subject \texttt{sheep}, the prediction is often on the central area because most of the ground truth locations for \texttt{sheep} related queries are close to the center of the image as shown in the statistics in Figure~\ref{fig:teaser}.
More examples are given in Figure~\ref{fig:1}, where such bias ubiquitously exists in not only nouns, but also verbs and prepositions.
For example, for \texttt{corner of} in Figure~\ref{fig:1}, 
the predicted region is biased to be of smaller size (than the size of the ground truth), since most of \texttt{corner of} samples in the dataset are smaller. The model simply takes this bias rather than justifies if the content in this region satisfies the query.

One may realize that the above absurd language bias is due to the \emph{confounder}, which is a common cause of other variables, rendering spurious correlations among them, even if they have no direct causal effects with each other~\cite{pearl2016causal}.  For example, the confounding effect of the \texttt{standing} bias in Figure~\ref{fig:1} may be due to the common sense that most standing people are the focus of the photographer. Thus, if we fail to consider such an effect that causes the correlation between \texttt{standing people} and the center location in training, when the confounder distribution changes in testing, \eg, most people standing aside, the former correlation in training will be no longer applicable in testing~\cite{pearl2014external}. Readers are encouraged to think about the confounders in other examples. In fact, similar confounding effect is also common in other classical vision tasks such as classification~\cite{yue2020interventional}, detection~\cite{tang2020long}, and segmentation~\cite{zhang2020causal}. However, compared to those single-modal tasks, the confounder in the visual grounding task is special. As this task is complex and multi-modal, the confounder is unobserved and non-enumerative in general. For example, a general ``common sense'' is elusive and hard to model. 

In this paper, we provide a theoretical ground that addresses why the visual grounding process is confounded by the unobserved confounder, and how we can overcome it. Our technical contributions are three-fold:

\begin{itemize}[leftmargin=+.05in]
\item To answer the above question, we frame the visual grounding pipeline into the causal graph. Thanks to the graph, we can model the causalities between language, vision, and locations, and thus offer the causal reasons why the grounding process is confounded. 

\item As the confounder is unobserved and non-enumerative, we propose a Referring Expression Deconfounder (RED) to estimate the substitute confounder based on the deconfounder theory~\cite{wang2019blessings}. In this way, we can mitigate the confounding bias without any knowledge on the confounder, \ie, by only using the same observational data just as other baseline methods of visual grounding.

\item We implement RED as a simple and effective language attention, whose embedding replaces the corresponding one in any grounding method. Therefore, RED is model-agnostic and thus can help various grounding methods achieve better performances.

\end{itemize}

\section{Related Work}
\noindent\textbf{Visual Grounding} 
In earlier works for visual grounding tasks, most of models are trained in two stages~\cite{liu2017referring,yang2019cross,chen2017query,plummer2018conditional,fang2019modularized,wang2019phrase}: its first stage detects region proposals using an off-the-shelf object detector~\cite{ren2015faster}, 
and
the second stage ranks these regions based on the distances between visual features and language embeddings.
Improved methods include
using a stronger attention mechanism~\cite{yu2018mattnet,deng2018visual,liu2019improving,yang2019dynamic,liu2019adaptive}, a better language parsing~\cite{liu2019learning,hong2019learning,niu2019variational,liu2020learning,dogan2019neural} or a better mechanism of filtering out poor region proposals~\cite{ref-nms}. 
While, it is often sub-optimal to train the model in separated stages. Most of
recent works~\cite{chen2018real,liao2020real,yang2019fast,sadhu2019zero} are thus proposing to exploit a one-stage (i.e., end-to-end) training paradigm.
Specifically, they fuse the image features and language embeddings in a dense and optimizable way, and directly predict the grounding result, i.e., the bounding box of the query object.
The improved methods~\cite{yang2020propagating,yang2020improving,shrestha2020magnet,lbyl} are mainly based on incorporating additional head networks that are used in two-stage models.

After careful comparison, we find that regardless of the different types of object detectors (e.g., YOLO and Faster R-CNN), the two-stage and one-stage methods do not have significant difference in terms of the learning pipeline, if all gradients can backpropagate through the backbone of the detector. Without loss of generality, this paper deploys multiple one-stage methods as baselines to show that our method is effective and generic.

\noindent\textbf{Causal Inference}.
Causal inference~\cite{rubin2019essential,pearl2016causal,bareinboim2012controlling,parascandolo2018learning,chiappa2019path} is often used to mitigate spurious correlations. Its methods include deconfounding~\cite{wang2020visual, zhang2020causal,yue2020interventional} and counterfactual inference~\cite{tang2020long,tang2020unbiased,niu2020counterfactual}.
The generic way is to disentangle the factors (in the task), and modeling the existing causal effects (among the factors) on the structural causal graph. 
In our work, we leverage the method of deconfounding to remove the bias we found in visual grounding datasets.
We understand that the confounder is often invisible and non-enumerative from the given dataset. We propose an approach to solve this by generating the proxy of the confounder inspired by a data generation method called Deconfounder~\cite{wang2019blessings}.

Our approach is different from existing deconfounding methods. For example, \cite{wang2020visual,yue2020interventional,zhang2020causal} select a specific substitute for confounder (like object class) and build the confounder set by enumerating its values.
\cite{qi2020two,yang2020deconfounded} use a learnable dictionary to learn the values of the confounder in the main training process.
In contrast, we build the dictionary by Deconfounder~\cite{wang2019blessings} first as an off-the-shelf part, and then in the grounding stage, we directly use the fixed deconfounder set based on which we are able to guarantees the confounder set is the cause of the feature but not vice versa.

\section{Visual Grounding in Causal Graphs}
\label{sec:3}
We frame the grounding pipeline summarized in Figure~\ref{fig:2} into a structural causal graph~\cite{pearl2016causal}, as shown in Figure~\ref{fig:3}, where the causalities among the ingredients, \eg, image, language query, and locations, can be formulated. 
The nodes in the graph denote data variables and the directed links denote the causalities between the nodes. Now, we introduce the graph notations and the rationale behind its construction at a high-level.
\subsection{Causal Graph}
\label{sec:3.1}

\noindent\bm{$X \rightarrow L \leftarrow R$}. $L$ is an object location. $X \rightarrow L$ and $R \rightarrow L$ denote that $L$ is directly detected from $X$ according to $R$.

\noindent\bm{$X \leftarrow G \rightarrow R$}. $G$ is the invisible (unobserved) confounder whose distribution is dataset-specific. For example, when $G$ is a dominant visual context (\eg, if there are more samples of \texttt{horses on grass} than \texttt{horse on road}, \texttt{on grass} dominates). $G\rightarrow X$ indicates that most images picturing \texttt{horse} and \texttt{grass}~\cite{zhang2020causal}, and $G\rightarrow R$ denotes that most language queries $R$ contain the words about \texttt{horse} and \texttt{grass}~\cite{zhang2018grounding}.

\noindent\bm{$G \rightarrow L$}. This is the ever-overlooked causation for the grounding: the confounder $G$ directly causes $L$, because the ground truth of $L$ is annotated under the same dataset containing the confounder $G$. For example, under the visual scene \texttt{horses on grass}, $G \rightarrow L$ indicates where and how to put $L$ (\eg, at the middle bottom position, with certain shape and size), which is the root of the bias in the grounding~\cite{zhang2020causal}.

\subsection{Causal Inference Process}
\label{sec:3.2}

\begin{figure}[t]
\centering
\includegraphics[width=2.5in]{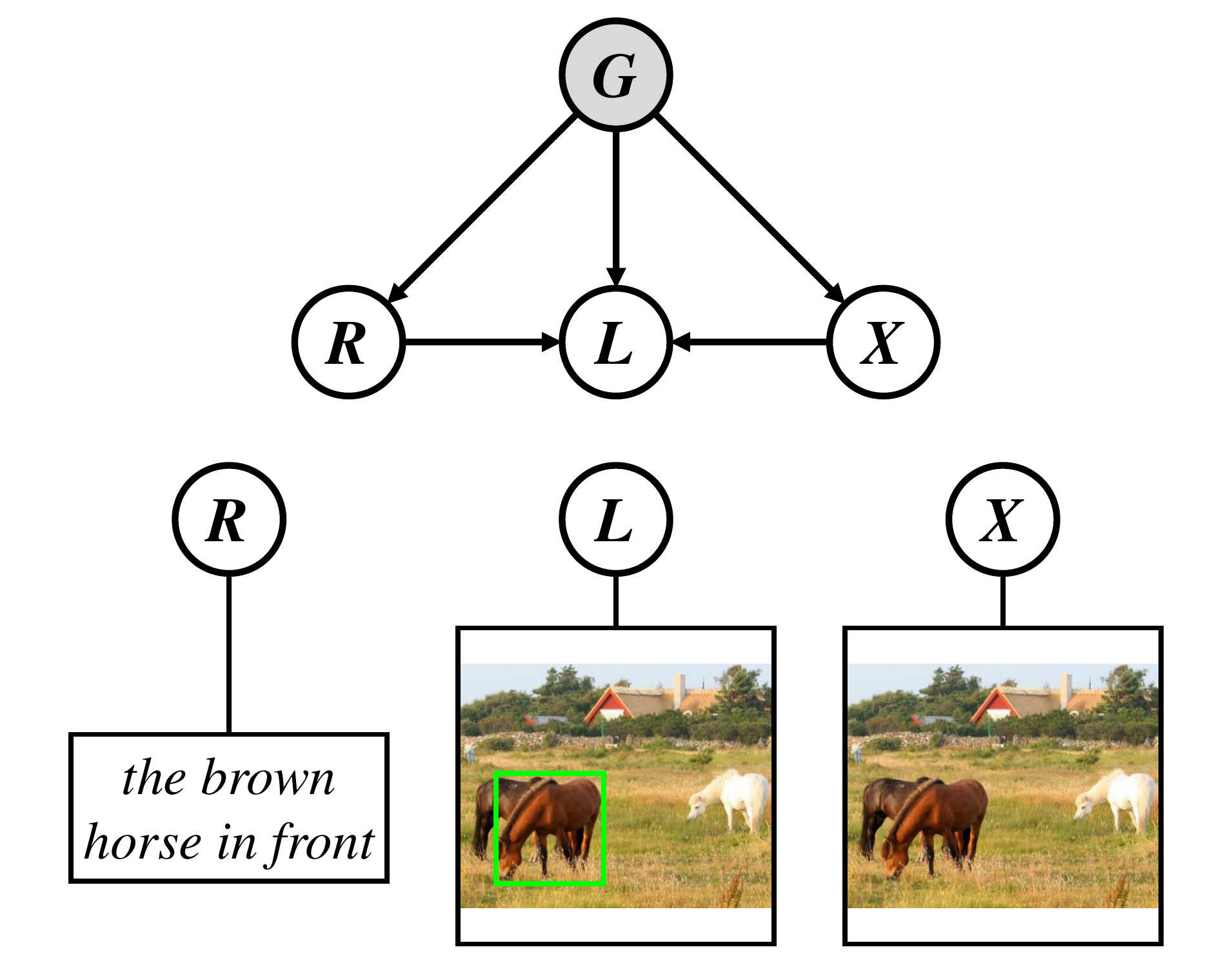}
\caption{The proposed causal graph and the examples of corresponding nodes. $G$: unobserved confounder, $X$: pixel-level image, $R$: language query, $L$: the location for the query.}
\label{fig:3}
\end{figure}

Given an image, the causal inference process of the grounding task is to maximize the intervention conditional probability of the object location $L$ according to the language query $R$, which is not equal to the observational conditional probability:
\begin{equation}
\centering
\label{eq:3.2.1}
  P(L\mid\textit{do}(R),X) \neq P(L\mid R,X),
\end{equation}
where the $do$-operation for $R$ denotes that we want to pursue the \emph{causal effect} of $R$ to $L$ by intervening $R=r$~\cite{pearl2016causal}, \ie, we expect that the model should predict location $L$ that varies from queries $R$. The inequality is because of the confounding effect~\cite{pearl2016causal}: Shown in Figure~\ref{fig:3}, even though we conditional on $X$ (\ie, cutting off the backdoor path $R \leftarrow G \rightarrow X \rightarrow L$), there is still another one, $R \leftarrow G \rightarrow L$, confounding $R$ and $L$. Therefore, this is the key bottleneck that the grounding pipeline is confounded by language. 

\noindent\textbf{Remarks}. One may argue that we should also intervene $X$, because $X$ leads another symmetric backdoor path like $R$ and $P(L|do(R), do(X))\!\neq\! P(L|do(R), X)$. Although this is true in general, we argue that such confounding effect does not only exist in the grounding task, but also ubiquitously exist in any visual recognition tasks~\cite{scholkopf2019causality}, which is not the main confounding effect in the language-dominant task.

To solve the confounding problem in visual grounding, the general way is to deconfound by using the backdoor adjustment~\cite{pearl2016causal} through controlling all the values of $G$:
\begin{equation}
\centering
\label{eq:3.2.2}
P(L\mid\textit{do}(R),X) = \mathbb{E}_{g\sim G}[P(L\mid R,X,g)].
\end{equation}
Yet, as the confounder is unobserved in general, we cannot enumerate its exact semantic, let alone control it.

\section{Deconfounded Visual Grounding}
\label{sec:4}

\subsection{Deconfounder}
\label{sec:4.1}

Thanks to Deconfounder~\cite{wang2019blessings}, we can derive a substitute confounder $\hat{G}$ from a data generation model $M$ even if the confounder is unobserved (invisible). Suppose we have the model $M$ that can generate $R$ from $\hat{G}$:
\begin{equation}
\centering
\label{eq:4.1.1}
P(R\mid\hat{G}) = \prod_{i=1}^{m}P(R_i\mid\hat{G}),
\end{equation}
where $R_i$ denotes the $i$-th feature of $R$ (\eg, the $i$-th dimension or $i$-th word embedding), respectively. $M$ ensures the independence among the features in $R$ conditional on $\hat{G}$, making $\hat{G}$ include all the confounder stratifications $g\sim G$. 
Here we give a sketch proof: If a confounder stratification (\ie, a value) $g$ is not included in $\hat{G}$, according to the causal graph illustrated in Figure~\ref{fig:3}, $g$ can simultaneously cause multiple dimensions in $X$, $R$ and $L$, making the different dimensions in $R$ dependent, which contradicts with Eq.~\eqref{eq:4.1.1}. Please kindly refer to the formal proof in Appendix.

Therefore, if we have a model $M$, we can sample $\hat{g}\sim \hat{G}$ to approximate stratifying the original $G$. By doing this, Eq.~\eqref{eq:3.2.2} can be re-written as:
\begin{equation}
\centering
\label{eq:4.1.2}
P(L\mid\textit{do}(R),X) = \mathbb{E}_{\hat{g}\sim \hat{G}}[P(L\mid R,X,\hat{g})].
\end{equation}
In practice, to avoid expensive network forward pass calculation, thanks to~\cite{xu2015show,yang2020auto}, we can apply the Normalized Weighted Geometric Mean to move the outer expectation into the feature level (The proof can be found in Appendix):
\begin{equation}
\centering
\label{eq:4.1.3}
\begin{split}
P(L|\textit{do}(R),X) &= \mathbb{E}_{\hat{g}\sim \hat{G}} [P(L| R,X,\hat{g})] \\
&\approx P(L\mid \mathbb{E}_{\hat{g}\sim \hat{G}}[R, \hat{g}] ,X).
\end{split}
\end{equation}
Next, we detail how to obtain the generative model $M$ and implement the above Eq.~\eqref{eq:4.1.3}.

\subsection{Referring Expression Deconfounder (RED)}
\label{sec:4.2}
Shown in the right part of Figure~\ref{fig:2}, our proposed approach includes two steps: training the generative model $M$ (illustrated as the shaded part) and training the deconfounded grounding model.

In the first step, we propose to use an auto-encoder to implement the generative model $M$. The auto-encoder encodes language query  $R$ into a hidden embedding, \ie, the substitute confounder $\hat{G}$ (denoted as pink bar), which can be decoded to the query $R'$:
\begin{equation}
\centering
\label{eq:4.2.1}
\begin{split}
& \textit{Encoder:} \quad \hat{G} = F_{enc}(R), \\
& \textit{Decoder:} \quad R' = F_{dec}(\hat{G}),
\end{split}
\end{equation}
where $F_{enc}$ and $F_{dec}$ are deterministic neural networks, \ie, once $R$ is set to specific values, the values of  $\hat{G}$ and $R'$ will be collapsed to a certain value with probability 1. Then, we can use a reconstruction loss $\mathcal{L}_{recon} = d(R,R')$ to train the generative model $M$, where $d(,)$ is a distance function for the features (\eg, we use Euclidean distance after feature pooling). Note that our deconfounder formulation also supports other non-deterministic generative models such as VAEs~\cite{kingma2013auto, nielsen2020survae} and conditional GANs~\cite{mirza2014conditional, douzas2018effective}.

After we derive the generative model $M$, we can collect all the samples of substitute confounders $\hat{g} \in \hat{G}$ for $R$, then cluster them to a dictionary $D_{\hat{g}}$ for $\hat{G}$ (denoted as the stack of pink bars in Figure~\ref{fig:2}). The reason of clustering is because the number of $\hat{g}$ is too large for backdoor adjustment, while a clustered dictionary can efficiently represent the main elements of $\hat{G}$.

In the second step, we use the dictionary $D_{\hat{g}}$ to perform deconfounded training by Eq.\eqref{eq:4.1.3}. We can instantiate the backdoor adjustment as:
\begin{equation}
\centering
\label{eq:4.2.2}
\mathbb{E}_{\hat{g}\sim \hat{G}}[R, \hat{g}] = \sum_{ \hat{g}\sim D_{\hat{g}}} f(r,\hat{g})P(\hat{g}),
\end{equation}
where the right-hand-side is summed over elements in the dictionary $D_{\hat{g}}$, $f(r,\hat{g})$ is a feature fusion function and $P(\hat{g})$ is $1/n$, which is an empirical prior. We implement $f(r,\hat{g})$ as a simple language attention mechanism and derive our deconfounded language feature $r'$ by:
\begin{equation}
\centering
\label{eq:4.2.4}
\begin{split}
& r' = \sum\nolimits_{ \hat{g}\sim D_{\hat{g}}} f(r,\hat{g})P(\hat{g}) \\
& = \sum\nolimits_{ \hat{g}\sim D_{\hat{g}}}Att(Q = r, K=\hat{g}, V = r)P(\hat{g}),
\end{split}
\end{equation}
where $Att(Q, K, V)$ can be any attention~\cite{vaswani2017attention} with the Query-Key-Value convention. In this paper, we only adopt a simple top-down attention. Recall in Figure~\ref{fig:2} to equip our RED with any conventional grounding method, all we need to do is just to replace the original language embedding feature $r$ (\ie, the light blue bar) with $r'$ (\ie, the purple bar).

Then, the overall deconfounded visual grounding model in Eq.~\eqref{eq:4.1.3} can be implemented as:
\begin{equation}
\centering
\label{eq:4.2.5}
\begin{split}
 &P(L = l\mid\textit{do}(R=r),X=x) \\
&\approx P(l\mid(\sum\nolimits_{\hat{g}} f(r,\hat{g})P(\hat{g}))\oplus x)=P(l\mid r'\oplus x),
\end{split}
\end{equation}
where $\oplus$ denotes feature concatenation and $l$ is a location index. In particular, the grounding model $P(l\mid\cdot)$ can be any detection head network that is a softmax layer outputing the probability of the $l$-th anchor box in the image~\cite{yang2019fast,yang2020improving}. Specifically, the reason why we only fuse $r$ and $\hat{g}$ without $x$ is because the ignorability principle~\cite{imai2004causal} should be imposed to the variable under causal intervention, that is, $R$ but not $X$. Table~\ref{table:3} demonstrates that other fusions that take $X$ as input will hurt the performance.

\begin{algorithm}[t]
	\caption{Visual Grounding with RED}
	\setstretch{1.15}
    \begingroup
    \LinesNumberedHidden
	\label{alg:1}
    \KwIn{Training images, language queries, and ground-truth locations $\mathcal{D} = \{(x_i,r_i,l_i)\}$} 
	\KwOut{Deconfounded grounding model $P_\theta(L=l\mid do(R=r), X=x)$ in Eq.~\eqref{eq:4.2.5}, whose illustration is in Figure~\ref{fig:2}}
	\Numberline Train the generative model $F_{enc}$ and $F_{dec}$ in Eq.~\eqref{eq:4.2.1};\\
	\Numberline Cluster the substitute confounders to derive $D_{\hat{g}}$;\\
	\Numberline Initialize grounding model parameters  $\theta$ randomly;\\
	\While{\textnormal{not converged}}{
	\Numberline Extract features ($x,r$) for sample $(x_i,r_i)$; \\
	\Numberline Calculate deconfounded language feature $r' = \sum_{\hat{g}} f(r,\hat{g})P(\hat{g})$ in Eq.~\eqref{eq:4.2.4};\\
	\Numberline Calculate deconfounded prediction $P(L=l\mid do(R=r), X=x)$ in Eq.~\eqref{eq:4.2.5};\\
	\Numberline Update $\theta$ by using the loss in Eq.~\eqref{eq:4.3};
	}
	\endgroup
\end{algorithm}

So far, we are ready to train the deconfounded grounding model with the following training losses:
\begin{equation}
\centering
\begin{split}
\label{eq:4.3}
\mathcal{L}_{overall} &= - \text{log} \, P(l_{gt}\mid r'\oplus x) +   \mathcal{L}_{reg}, \\
\end{split}
\end{equation}
where $l_{gt}$ is the ground-truth location and $\mathcal{L}_{reg}$ is a regression loss for the ground-truth bounding box~\cite{yang2019fast,yolov3} or mask~\cite{luo2020multi}. The whole implementation is summarized in Algorithm~\ref{alg:1}.

\section{Experiments}
\subsection{Datasets}
\label{sec:5}

To evaluate RED, we conducted extensive experiments on the following benchmarks.
\textbf{RefCOCO}, \textbf{RefCOCO+} and \textbf{RefCOCOg} are three visual grounding benchmarks and their images are from MS-COCO~\cite{mscoco}. 
Each bounding box from MS-COCO object detection ground truth is annotated with several referring expressions.
RefCOCO~\cite{refcoco} has 50,000 bounding boxes with 142,210 referring expressions in 19,994 images and is split into train$/$ validation$/$ testA$/$ testB with 120,624$/$ 10,834$/$ 5,657$/$ 5,095 images, respectively. RefCOCO+~\cite{refcocog} has 19,992 images with 141,564 referring expressions and is split into train$/$ validation$/$ testA$/$ testB with 120,191$/$ 10,758$/$ 5,726$/$ 4,889 images, respectively. 
Note that, during testing, RefCOCO and RefCOCO+ provide two splits as testA and testB. 
Images in testA contain multiple people and images in testB contain multiple instances of all other objects. 
For RefCOCOg, we deploy the popular data partition called RefCOCOg-umd~\cite{refcocog-umd}, and denote the partitions as val-u and test-u in Table~\ref{table:1}. We also conduct experiments on \textbf{ReferItGame}~\cite{referit} and \textbf{Flickr30K Entities}~\cite{flickr30k-entity}. More details can be found in the supplementary materials.

\begin{table*}[ht]
\centering
\setlength\tabcolsep{1.0pt}
\scalebox{0.85}{
\renewcommand\arraystretch{1.05}
\begin{tabular}{p{0.6cm}<{\centering}p{0.6cm}<{\centering}|l|p{0.2cm}<{\centering} lll lll ll}
\toprule
&&
\multirow{2}{*}{Methods} &&
  \multicolumn{3}{c}{RefCOCO} &
  \multicolumn{3}{c}{RefCOCO+} &
  \multicolumn{2}{c}{RefCOCOg} \\  \cline{4-12} 
  ~ & ~ &
   && val & testA & testB & val & testA & testB & val-u &
  test-u \\ \hline 
\multirow{5}{*}{\rotatebox{90}{Two-Stage}}& \multirow{5}{*}{\rotatebox{90}{SOTA}}
&MattNet\cite{yu2018mattnet}  && 76.40  & 80.43 & 69.28&64.93& 70.26& 56.00  & 66.67& 67.01\\
&&NMT\cite{liu2019learning}     & & 76.41   & 81.21   & 70.09  & 66.46 & 72.02 & 57.52    & 65.87   & 66.44    \\
&&CM-Att\cite{liu2019improving}   && 78.35   & 83.14   & 71.32 & 68.09 & 73.65 & 58.03   & 67.99   & 68.67    \\
&&DGA (Yang et al. 2019)   && -  & 78.42   & 65.53 & - & 69.07  & 51.99        & - & 63.28    \\ 
&&Ref-NMS~\cite{ref-nms}   && 80.70  & \textbf{84.0}   & 76.04 & 68.25 & 73.68  & 59.42 & 70.55 & 70.62    \\ \cmidrule{1-12}\morecmidrules
\multirow{12}{*}{\rotatebox{90}{One-Stage}}&
\multirow{8}{*}{\rotatebox{90}{SOTA}}
&RCCF~\cite{liao2020real} & &-  & 81.06   & 71.85   & - & 70.35   & 56.32  & -    & 65.73   \\
&~&Yang's-V1~\cite{yang2019fast} && 72.54   & 74.35   & 68.50           & 56.81 & 60.23   & 49.60    & 61.33   & 60.36\\
&~&Iterative (Sun el al. 2021) & &-  & 74.27   & 68.10   & - & 71.05   & 58.25  & -    & 70.05   \\
&~&PFOS~\cite{pfos} & &79.50  & 81.49   & 77.13   & 65.76 & 69.61   & 60.30  & 69.06    & 68.34   \\
&~&LBYL~\cite{lbyl} & &79.67  & 82.91   & 74.15   & 68.64 & 73.38   & 59.49  & -    & -   \\
&~&Yang's V1$^\dagger$ (YOLO-V4) & & 74.67  & 75.98   & 70.76   & 57.21 & 61.11   & 50.93  & 61.89    & 61.56   \\
& &
{MCN~\cite{luo2020multi}} && 
{80.08}&
{82.29}&
{74.98}&
{67.16}&
{72.86}&
{57.31}&
{66.46}&
{66.01}  \\
 & ~& MCN (fixed BERT)  && 78.93 &82.21 & 73.95&65.54 & 71.29 & 56.33 &65.76&66.20 \\
\cmidrule{2-12}\morecmidrules
&\multirow{4}{*}{\rotatebox{90}{OURS}}
&\makecell[l]{{Yang's-V1 +RED}} &\cellcolor{llgray}&
  \cellcolor{llgray}78.04\scriptsize\color{red}+5.50 &
  \cellcolor{llgray}79.84\scriptsize\color{red}+5.49 &
  \cellcolor{llgray}74.58\scriptsize\color{red}+6.08 &
  \cellcolor{llgray}62.82\scriptsize\color{red}+6.01 &
  \cellcolor{llgray}66.03\scriptsize\color{red}+5.80 &
  \cellcolor{llgray}56.72\scriptsize\color{red}+7.12 &
  \cellcolor{llgray}66.40\scriptsize\color{red}+5.07 &
  \cellcolor{llgray}66.64\scriptsize\color{red}+6.28 \\
&~&\makecell[l]{{Yang's-V1$^\dagger$+RED}}& \cellcolor{llgray}&
  \cellcolor{llgray}78.76\scriptsize\color{red}+4.09 &
  \cellcolor{llgray}80.75\scriptsize\color{red}+4.77 &
  \cellcolor{llgray}76.19\scriptsize\color{red}+5.43 &
  \cellcolor{llgray}63.85\scriptsize\color{red}+6.64 &
  \cellcolor{llgray}66.87\scriptsize\color{red}+5.76 &
  \cellcolor{llgray}57.51\scriptsize\color{red}+6.58 &
  \cellcolor{llgray}69.46\scriptsize\color{red}+7.57 &
  \cellcolor{llgray}69.51\scriptsize\color{red}+7.95 \\

 & ~&\makecell[l]{{MCN (fixed BERT) +RED}}& \cellcolor{llgray}
  & \cellcolor{llgray}79.23\scriptsize\color{red}+0.30
  & \cellcolor{llgray}83.22\scriptsize\color{red}+1.01
  & \cellcolor{llgray}75.23\scriptsize\color{red}+1.28
  & \cellcolor{llgray}66.84\scriptsize\color{red}+1.30
  & \cellcolor{llgray}72.47\scriptsize\color{red}+1.18
  & \cellcolor{llgray}59.03\scriptsize\color{red}+2.70
  & \cellcolor{llgray}67.28\scriptsize\color{red}+1.52
  & \cellcolor{llgray}67.02\scriptsize\color{red}+0.82 \\
  
&~&
  \makecell[l]{{LBYL +RED}}& \cellcolor{llgray}&
  \cellcolor{llgray}\textbf{80.97}\scriptsize\color{red}+1.30 &
  \cellcolor{llgray}83.20\scriptsize\color{red}+0.29 &
  \cellcolor{llgray}\textbf{77.66}\scriptsize\color{red}+3.51 &
  \cellcolor{llgray}\textbf{69.48}\scriptsize\color{red}+0.84 &
  \cellcolor{llgray}\textbf{73.80}\scriptsize\color{red}+0.42 &
  \cellcolor{llgray}\textbf{62.20}\scriptsize\color{red}+2.71 &
  \cellcolor{llgray}\textbf{71.11} &
  \cellcolor{llgray}\textbf{70.67} \\
  \bottomrule
\end{tabular}}
\caption{The performance (Acc@0.5\%) compared to the state-of-the-art (SOTA) methods on RefCOCO, RefCOCO+ and RefCOCOg, where 
$^\dagger$ denotes applying the visual feature backbone from YOLO-V4, methods with citation denote that the results are from the cited papers. Bold text denotes the best performance. We will use the same notation in the following tables. Note that our reproduced performance of MCN is different from the original paper, because we applied fixed BERT instead of LSTM for language embedding and the reason can be found in ``\textbf{Implementation Details}''.}
\label{table:1}
\end{table*}

\begin{table}[t]
\centering
\renewcommand\arraystretch{1.1}
\setlength\tabcolsep{2.0pt} 
\scalebox{0.8}{
\begin{tabular}{l|c|c}
\toprule
\multicolumn{1}{c|}{{Method}} &{\centering ReferIt Game} & 
{\centering Flickr30K Entities}

 \\ \hline
ZSGNet (Sadhu el al. 2019) & 58.63 & 63.39 \\
RCCF~\cite{liao2020real}  & 63.79 & - \\
Yang's-V1~\cite{yang2019fast} & 60.67 & 68.71 \\
Yang's-V2~\cite{yang2020improving} & 64.33 & 69.04\\
LBYL~\cite{lbyl} & 66.51 & - \\ 
Yang's-V1$^\dagger$ (YOLO-V4) & 61.21& 70.25 \\
\hline

\makecell[l]{{Yang's-V1$^\dagger$ +RED}} & \cellcolor{llgray}64.37\scriptsize\color{red}+3.16 &  \cellcolor{llgray}\textbf{70.50}\scriptsize\color{red}+0.25 \\
\makecell[l]{{Yang's-V2 +RED}} &\cellcolor{llgray}66.09\scriptsize\color{red}+1.76 &  \cellcolor{llgray}69.76\scriptsize\color{red}+0.72 \\
\makecell[l]{{LBYL +RED}} &\cellcolor{llgray}\textbf{67.27}\scriptsize\color{red}+0.76 &\cellcolor{llgray}  -\\ \bottomrule
\end{tabular}}
\caption{Comparing with SOTA methods on the test sets of ReferItGame and Flickr30K Entities (Acc@0.5\%). It is reasonable that ours gets significantly higher gains on the more difficult dataset---ReferIt Game. Please refer to ``\textbf{Comparing with the state-of-the-art (SOTA)}''.}
\label{table:2}
\end{table}

\subsection{Implementation Details}
As our RED is model-agnostic, We followed the same settings of implemented methods to extract the visual representations. 
As for language representations, before the grounding model training, we used the embedding extracted from uncased version of BERT~\cite{bert} to train an auto-encoder.
Then, we used the trained auto-encoder to extract the substitute confounders for all the training samples $R$. We deployed the K-Means algorithm to cluster those into $N=10$ clusters forming the confounder dictionary $D_{g^*}$ in Eq.~\eqref{eq:4.2.2}. 
In the grounding model training stage, we first used a pre-trained frozen BERT to extract language embeddings from the query.  Note that we applied the same frozen BERT used in auto-encoder training to make the extracted embeddings consistent with our dictionary.
Second, we computed the deconfounded language embeddings by Eq.~\eqref{eq:4.2.4}. Then, it will be concatenated to the visual features as shown in Figure~\ref{fig:2}.
Limited by the theory of deconfounder, to ensure the validity of confounder embeddings, we have to use the same structure (\ie, the same frozen BERT structure) to prevent the embeddings gap between the substitute confounders and deconfounded training. As a finetuned BERT is more popular, We will closely follow the breakthroughs of the deconfounder to improve this implementation.
More other details of implementations and training settings can be found in the supplementary materials.

\begin{table}[t]
\centering
\scalebox{0.85}{
\renewcommand\arraystretch{1.1}
\begin{tabular}{l|cc|cc}
\toprule
\multicolumn{1}{c|}{\multirow{2}{*}{Methods}} &
  \multicolumn{2}{c|}{RefCOCO+} &
  \multicolumn{2}{c}{RefCOCOg} \\ \cline{2-5} 

\multicolumn{1}{c|}{} &testA &testB &val-u &test-u \\ \hline
 Yang's-V1$^\dagger$ &61.11 &50.93 &61.89 &61.56         \\\hline
\makecell[l]{{+\textrm{AE($X\oplus R$,$X'\oplus R'$)}}}   &66.74 &57.23 & 69.01  & 68.55  \\
\makecell[l]{{+\textrm{AE($R$,$R'$)} (our RED)}}  &\textbf{66.87} &\textbf{57.51}  & \textbf{69.46} & \textbf{69.51} \\\hline
\makecell[l]{{+\textrm{Att($X$)}+\textrm{Att($R$)}}}  &65.57 &56.37 & 67.83  & 68.11  \\
\makecell[l]{{+\textrm{Att($R$)} (our RED)}}     &\textbf{66.87} &\textbf{57.51}  & \textbf{69.46} & \textbf{69.51}   \\

\bottomrule
\end{tabular}}
\caption{The ablation study of using different auto-encoders and different fusion methods. $X$ and $R$ denote image and language query, respectively.
\textrm{AE(input value, output value)} denotes the auto-encoder in Eq.~\eqref{eq:4.2.1} fed with different input values for generating corresponding output values. \textrm{Att(value)} denotes the attention operation in Eq.~\eqref{eq:4.2.4} using different values of $V$.
}
\label{table:3}
\end{table}

\subsection{Quantitative Results}
\label{sec:5.3}
\noindent\textbf{Comparing with the state-of-the-art (SOTA)}. In Table~\ref{table:1}, we summarize the results of SOTA methods and those \emph{w/} and \emph{w/o} our RED, on RefCOCO-series datasets. We used 4 SOTA methods as our baselines: Yang's-V1~\cite{yang2019fast}, Yang's-V2~\cite{yang2020improving}, LBYL~\cite{lbyl}, and MCN~\cite{luo2020multi}.
Table~\ref{table:2} shows the results on the ReferItGame and Flickr30K Entities. %
From an overall view, we observe that using our RED method, which means mitigating the bias in language features, improves the performance of grounding models generally.
We elaborate the improvements in the following two points.

\noindent\textbf{RED achieved more improvements for RefCOCOg.} 
This is because RefCOCOg have longer queries than other datasets.
Our approach is used to mitigate the bias in language queries. Longer queries contain more bias, so they can benefit more from our approach.
For example, our gain on ReferItGame is more than that on Flickr30K Entities, \ie, an increase of 3.16\% vs. 0.25\%, as most of queries in Flickr30K Entities (\eg, \emph{a man}, \emph{a bench}) are as short as class labels, which contain very little language bias.

In addition, we find that RED can improve more on testB than on testA or val on RefCOCO and RefCOCO+. The reason is that the distribution of testB data is more discrepant than that of training data. It means the testing on testB suffers more from the bias in the grounding model. 

\begin{figure}[t]
\centering
\includegraphics[width=3.0in]{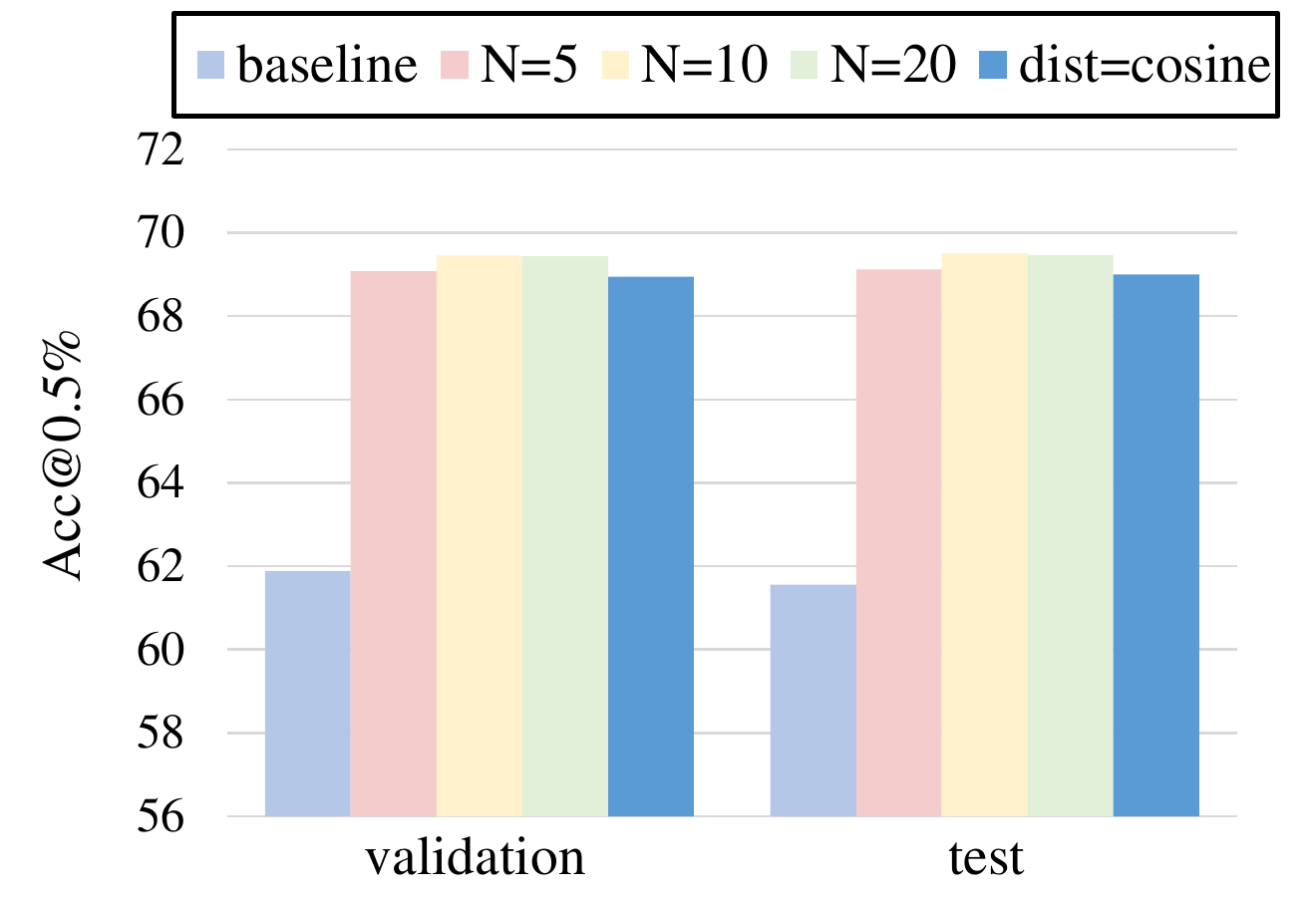}
\caption{Results using different values of cluster number $N$ and distances for clustering from RefCOCOg.
Note that the skyblue, pink, yellow and green columns use Euclidean distance (default), and the blue with $N=10$ (default).}
\label{fig:5}
\end{figure}

\noindent\textbf{RED improved Yang's-V1 (or -V1$^\dagger$) the most}. 
The improvement of RED is brought by applying the deconfounding operation---$do(R)$ (and optionally using $do(X)$).
While, those methods with complex co-attention operations and additional visual controlling, e.g., MCN and LBYL, may downplay the effectiveness of deconfounding.
In specific, MCN uses extra mask supervision to enforce model looking at the shapes rather than only the bounding boxes. Because language confounding effect only influences the location $L$, the additional shape in the objective function will not be confounded. Therefore, the whole training process of MCN will take less language confounding effects. LBYL splits the image features into four parts combining with queries respectively. This is to intervene $X$ by the value of each part of $X$, playing the similar role of $do(X)$. Therefore, plugging our RED brings only a partial margin of improvement.

\noindent\textbf{Justification for $do(R)$ and attention for $R$}. 
We use AE($X\oplus R$,$X'\oplus R'$) to evaluate the performance of language and vision deconfounder for $do(R, X)$, where $\oplus$ denotes concatenation operation. Specifically, before training grounding models, we reconstruct the embeddings of $R$ and $X$ simultaneously, and drive the dictionary by clustering the language and vision confounder using the same way in ``\textbf{Implementation Details}''. As shown in Table~\ref{table:3}, we find RED can achieve higher improvements than language and vision deconfounder. The reason may be the embeddings of $X$ need to change during the grounding training, which hurts the consistency of the dictionary of vision deconfounder. Then, we perform attention on both $X,R$ (\ie,$+Att(X)+Att(R)$) by the substitute confounder, where the attention for $X$ is implemented as a normal channel attention following SENet.  We found the performance drop in this ablation. This generally conforms to our statement for Eq~\eqref{eq:4.2.5}: To pursue the causal effect from $R$ to $L$, we need to use the confounder of $R$ and conduct intervention on $R$.

\noindent\textbf{Hyperparameter selection.} We explore different numbers of clustering centers and two clustering distance metrics. As shown in Figure~\ref{fig:5}, all the clustering numbers $N$ from 5 to 20 show significant improvements over the baseline. After $N$ exceeding 10, the performance won't show further improvement, thus we set $N=10$. In the comparison of clustering criterion, Euclidean distance shows constant superiority, which is reasonable because of the usage of $L_2$ loss in the auto-encoder.

\noindent\textbf{Time Consumption}. As a significant advantage of the one-stage grounding model is its speed, we test the overhead of our RED in the inference stage to ensure it will not hurt the speed. Under fair settings, we test the speed of Yang's-V1 and Yang's-V1+RED on a single Tesla V100. The results are 24.56ms and 26.53ms per sample, respectively, and the overhead is only 8\%, which is reasonable.

\subsection{Qualitative Results}
\label{sec:5.4}

\begin{figure}[t]
\centering
\includegraphics[width=3.2in]{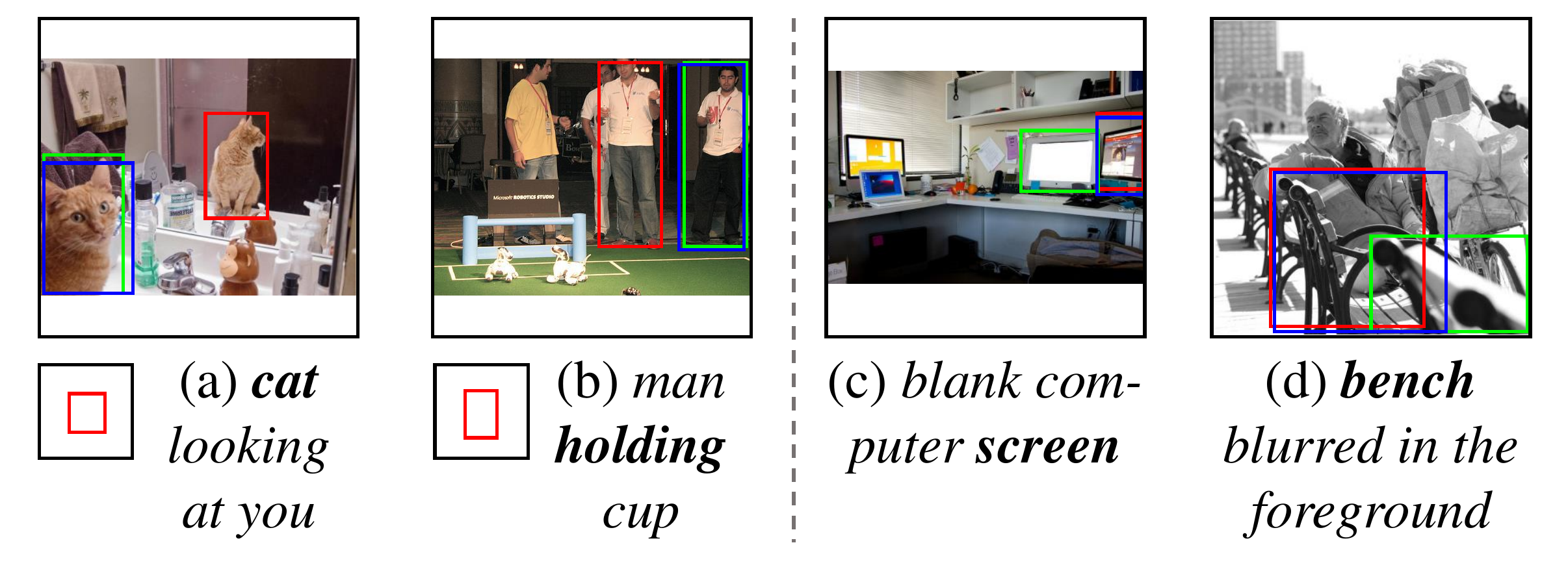}
\caption{Qualitative examples. 
The left two examples show that RED mitigates the language bias in the results of Yang’s-V1$^\dagger$, where the red boxes denote the biased predictions of Yang’s-V1$^\dagger$, blue boxes denote prediction of Yang’s-V1$^\dagger$+RED and greens ones are ground-truth boxes. The right two examples show our failure cases.
}
\label{fig:4}
\end{figure}

\noindent\textbf{Language Bias Corrections}. The qualitative results illustrated in the left half in Figure~\ref{fig:4} show that our RED can help models to remove the spurious correlations. In (a) and (b), the location bias is drawn at the left-bottom corner which misleads the grounding models to choose red boxes. After applying our RED, we find that the model can counter such bias (\ie, predict blue boxes near to the ground truth).

\noindent\textbf{Failure Cases}. We still find some failure cases shown in the right in Figure~\ref{fig:4}. In (d), for example, both the red and blue boxes (predicted by baseline and +RED) tend to select the colorful computer \texttt{screen}, which is more common in the dataset. On the contrary, \texttt{bench} is the salient subject compared to the complicated description. That's why both models tend to choose the middle bench.

\section{Conclusions}
We investigated the confounding effect that exists ubiquitously in visual grounding models, where the conventional methods may learn spurious correlations between certain language pattern and object location. As the confounder is unobserved in general, we proposed a Referring Expression Deconfounder (RED) approach that seeks a substitute one from observational data. The implementation result of RED is just a simple language attention feature that replaces the language embeddings used in any grounding method. By doing this, RED improves various strong baselines consistently. Our future plans include: 1) We will use other generative models to implement RED~\cite{nielsen2020survae}.  2) As hacking the unobserved confounder is ill-posed in general~\cite{d2019multi}, we will try to introduce the large-scale vision-language pre-training priors~\cite{lu202012, lu2019vilbert} into the deconfounder. 

{\small
\bibliography{aaai22}
}


\section*{Supplementary material}

This supplementary material will provide further details for the main paper, including A. Detailed basic knowledge of causal graph, B. Proofs and details which are omitted in the main paper, C. More qualitative examples to justify the effectiveness of our RED.

## Basic Knowledge of Causal Graph

### Causal Graph

**This subsection is supplemented for Section "Causal Graph" in the main paper.** Causal graph (Pearl, Glymour, and Jewell 2016) is a high-level road-map indicating how the variables interact with each other (i.e., revealing the causal relationships between them). In general, it is illustrated by a directed acyclic graph $\mathcal{G} = \{\mathcal{N}, \mathcal{E}\}$, containing nodes and arrows, where nodes $\mathcal{N}$ represent variables and arrows $\mathcal{E}$ (i.e., directed links) represent causal relationships between them. For example, Figure 1(a), one of the basic causal graph structures, which we will introduce later, contains $X \to Z$ which represents that $X$ causes $Z$ or the values of $Z$ listen to $X$, while $X \not\to Y$ represents there is no **direct** causal effect from $X$ to $Y$.

There are 3 basic structures of causal graph: **Chain**, **Fork** and **Collider**. Here, we show the first two in Figure 1, as our paper does not include the collider, and it is recommended to read the book (Pearl, Glymour, and Jewell 2016) for more details. **Chain** describes that, for the middle node $Z$, there is one arrow directed into it (i.e., $X \to Z$), and one arrow directed out of it (i.e., $Z \to Y$). **Fork** describes that, for the middle node $Z$, there are two arrows emanated from it (i.e., $Z \to X$ and $Z \to Y$).

In these structures, there are several straightforward dependencies: 1). The two variables with a directed link are dependent (e.g., $X$ and $Z$ are dependent in Figure 1(a), as $P(Z = z|X = x) \neq P(Z = z)$). 2). The two variables without a directed link can be conditional independent (e.g., in Figure 1(b), $X$ and $Y$ are independent, conditional on $Z$, as $P(Y = y|Z = z, X = x) = P(Y = y|Z = z)$). Here, we provide an intuitive example for a better understanding of conditional independence: We use $X$ to represent "sales of ice cream", $Y$ to represent "crime ratio" and $Z$ to represent "temperature" in Figure 1(b), respectively. The links are obvious that with the increase of "temperature", "sales of ice cream" ($X$) will grow and "crime ratio" ($Y$) will also grow (e.g., harassment in summer). If we condition on $Z$, like to observe the data of $X$ and $Y$ in summer (i.e., certain $Z$), there will be no relationship between them as they irregularly float at a high level. Otherwise, we will find the dependence between $X$ and $Y$ as they grow and decrease together in the whole year.

### Confounder

**This subsection is supplemented for Section "Causal Graph" in the main paper.** Illustrated in Figure 1(b), the fork structure contains a **confounder** $Z$, which both causes $X$ and $Y$, and opens a non-causal path (i.e., backdoor) started from $X$ (i.e., $X \leftarrow Z \to Y$), making them spuriously correlated with each other. Use the previous example, the "temperature" makes the "sales of ice cream" $X$ and "crime

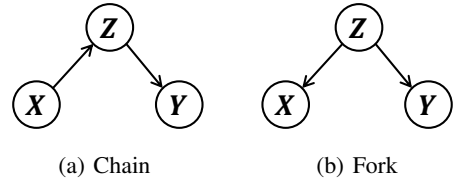

Figure 1: Two basic structures of causal graph.

ratio" $Y$ have dependence (i.e., they jointly grow and decrease in the whole year). But, such dependence cannot represent the causal effect because we know the cause of $Y$ is not $X$ (i.e., the causal effect from $X$ to $Y$ is 0).

In general, we use $P(Y|X) - P(Y)$ to represent the effect of $X$ on $Y$ (Note that as $P(Y)$ is a prior, we usually only calculate $P(Y|X)$), and for naive causal graph $X \to Y$, as there is no other path between them, it is also the causal effect that we want to pursue. However, in the "ice-crime" example, we find that $P(Y|X) - P(Y)$ is not zero, as their dependence makes $P(Y|X) \neq P(Y)$, which cannot represent the causal effect of $X$ on $Y$ (i.e., 0) anymore. As a result, we need new notations to formulate the causal effect under the confounder.

### Do and Backdoor Adjustment

**This subsection is supplemented for Section "Causal Inference Process" in the main paper.** Thanks to the $do$ factor introduced in (Pearl, Glymour, and Jewell 2016), we can use $P(Y|do(X = x)) - P(Y)$ to formulate the causal effect of $X$ on $Y$, where $do(X = x)$ denotes that we set a value $x$ to $X$ rather than observing $X = x$ (For convenience, we use $do(x)$ to denote $do(X = x)$ in the following parts). The reason is that when we $do$ a variable, we cut off all the paths incoming to the variable, which is equal to cut all the backdoors, ensuring that all the effects calculated are from the causal path. In the previous example, $do(x)$ denotes that we manually set the "sale of ice cream" to the value $x$ and then observe the change of "crime ratio". Of course, there will be no change as they do not have any causal relationships, indicating that the result for $P(Y|do(x)) - P(Y)$ is 0, which conforms to our commonsense. After we provide some preliminaries, we will give the theoretical proof for the example.

Now, we want to calculate the $do$ formula by the observational data, because we cannot really set a variable to some values and observe the results in the fixed dataset. Therefore, we need to transform the $do$ formula to the general probability formula. Thanks to the $do$-calculus (Pearl, Glymour, and Jewell 2016), there are 3 rules for eliminating the $do$-factor to derive the probability form.

**Rule 1.** If there is no causal path from $X$ to $Y$ (i.e., no path which starts from $X$, maybe mediated by some nodes and finally ends at $Y$), we can directly remove $do(x)$:

$$P(Y|do(x)) = P(Y). \quad (1)$$

**Rule 2.** When $X$ and $Y$ are independent conditional on

$Z$, the probability for $Y$ will not be influenced by $X$:

$$P(Y|z, X) = P(Y|z). \quad (2)$$

**Rule 3.** If all backdoors from $X$ to $Y$ are blocked by a set of variables $Z$, then $do(x)$ is equivalent to $observe(x)$:

$$P(Y|do(x), z) = P(Y|x, z). \quad (3)$$

Now, for the "ice-crime" example, we can use these rules to perform backdoor adjustment to eliminate $do$ in the causal effect formula (i.e., $P(Y|do(x)) - P(Y)$), by adjusting the confounder $Z$:

$$\begin{aligned} &P(Y|do(x)) - P(Y) \\ &= \sum_z P(Y|do(x), z) P(z|do(x)) - P(Y) \\ &= \sum_z P(Y|x, z) P(z) - P(Y) \quad (4) \\ &= \sum_z P(Y|z) P(z) - P(Y) \\ &= P(Y) - P(Y) = 0, \end{aligned}$$

where the first derivation using basic Bayes rules, second line using **Rule 3** (i.e., $Z$ can cut the backdoor path $X \leftarrow Z \rightarrow Y$) and **Rule 1** (i.e., there is no causal path from $X$ to $Z$), the third one using **Rule 2** (i.e., conditional on $Z$, $X$ and $Y$ are independent), and the last one is obvious. We find that the result of our causal effect formula is 0, which again conforms to our commonsense. In general, backdoor adjustment is represented as (i.e., the second line):

$$P(Y|do(x)) = \mathbb{E}_{z \sim Z}[P(Y \mid x, z)], \quad (5)$$

which is equivalent to Equation 3 in the main paper. In conclusion, we use $do(x)$ to denote the causal effect, when $P(Y|X)$ cannot denote it as the existence of confounder, and then we can use backdoor adjustment to control the confounders which can cut off all the backdoors to calculate the $do$ formula by observational data (i.e., probability formula). However, as we mentioned, if the confounder set is unobserved, it is impossible to control all the confounders. Then, we need some substitutes to perform backdoor adjustment.

Note that, after we introduce the basic knowledge of causal graph and backdoor adjustment, we will combine the situation in the main paper to introduce the following parts, where the notation of confounder is $G$ instead of $Z$, causes are $R$ instead of $X$ and potential outcome is $L$ instead of $Y$. As we conditional on the images, we only illustrate how to perform backdoor adjustment on the path $R \leftarrow G \rightarrow L$.

### Ignorability and Deconfounder

**This subsection is supplemented for Section "Deconfounder" in the main paper. Ignorability.** There is an assumption, ignorability (Rosenbaum and Rubin 1983; Imai and Van Dyk 2004), if we want to safely use a set of $G$ to perform the backdoor adjustment. It says that if the treatment assignment is independent to potential outcomes conditional on the set $G$, then it is valid to use $G$ to perform backdoor adjustment:

$$A_r \perp\!\!\!\perp L(r)|G, \quad (6)$$

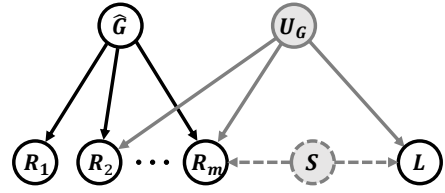

Figure 2: A graphical model to reason about conditional dependencies, which is not a causal graph, where $R_i$ denotes $i$-th dimension of $R$, $L$ denotes the potential outcome, $\hat{G}$ denotes substitute confounder, $U_G$ denotes assumed multi-cause confounder not in $\hat{G}$ and $S$ denotes single-cause confounder which does not exist here by the assumption.

where $A_r$ is the assignment for causes $R$, which can be understood as how to select $R$ in the data generation stage (i.e., treatment assignment) and $L(r)$ is the potential outcome. For example, if we want to test the effectiveness of a new drug and advertise volunteer recruitment in the city, then we will more likely to choose urban people as treatment individuals because they have more chance to notice the advertisement. This underlying selection makes the treatment assignment dependent on the outcome as the constitution of urban people is different from the rural ones, which is another factor influencing the outcome.

**Deconfounder.** Under the situation that the confounder $G$ is unobserved, if we can provide a substitute $\hat{G}$ which satisfies ignorability, we can still perform causal inference by backdoor adjustment on $\hat{G}$. Thanks to the Deconfounder (Wang and Blei 2019), it provides a way to approximate multi-cause confounder (illustrated in Figure 2), which satisfies ignorability under a weak assumption (no single-cause confounder). The steps of building Deconfounder are introduced in the main paper: 1) Build a generative model $M$ from $\hat{G}$ to causes $R_i, i = 1, 2, ..., n_R$. 2) Perform backdoor adjustment on $\hat{G}$ or its approximation $D_{\hat{g}}$. In the next section, we will give details about why the Deconfounder satisfies ignorability.

## Proofs and Details

### Detailed Proof for Deconfounder

**This subsection is supplemented for the detailed proof of "$\hat{G}$ include all the confounder stratifications $g \sim G$" in the Section "Deconfounder" in the main paper.** $d$-**separation**. For the following proofs, we first introduce a lemma indicating when the independence holds (Pearl, Glymour, and Jewell 2016): A path $p$ is blocked by a set $Z$ if and only if 1). $p$ contains a chain of nodes $A \rightarrow B \rightarrow C$ or a fork $A \leftarrow B \rightarrow C$ such that the middle node $B$ is in $Z$ (i.e., $B$ is conditioned on), or 2). $p$ contains a collider $A \rightarrow B \leftarrow C$ such that the collision node $B$ is not in $Z$ (Note that the second one will not be used, just for completeness).

According to the definition of our data generation model,

| Index | Input | Operation | Output |
|-------|-------|-----------|--------|
| (1) | R(1×20) | Bert | r(20×768) |
| (2) | r | Average | $\bar{r}$(1× 768) |
| (3) | $\bar{r}$ | Linear/ReLU | $\hat{g}$(1×64) |
| (4) | $\hat{g}$ | Linear/ReLU | $\bar{r}'$ (1×768) |

Table 1: The implementation details of our deconfounder, where $\hat{g}$ denotes the substitute confounder feature, $\bar{r}$ denotes the average of language features.

we write the Equation 3 in the main paper at here:

$$P(R_1, R_2, ..., R_m \mid \hat{G}) = \prod_{i=1}^{m} P(R_i \mid \hat{G}). \quad (7)$$

That means conditional on $\hat{G}$, the independence between $R_i$ holds. Illustrated in Figure 2, if we assume that there exists an unobserved multi-cause confounder $U_G$, which links to multiple causes $R_i$ and the outcome $L$, then the causes will be dependent according to the $d$-**separation** (i.e., $R_i$ connects to each other by $U_G$, even if we conditional on $\hat{G}$, as $\hat{G}$ does not include all the middle points (i.e., $U_G$) according to the assumption), which contradict to Eq (7). Therefore, such multi-cause confounder $U_G$ cannot exist.

As we further assume that there is no single-cause confounder, illustrated as $S$ in Figure 2, the independence statement implies that there is no backdoor from $R$ to $L$ when we condition on $\hat{G}$ (i.e., block all backdoors through $\hat{G}$), which implies ignorability (i.e., $A_r \perp\!\!\!\perp L(r)|\hat{G}$), justifying the causal inference by using $\hat{G}$ to perform the backdoor adjustment.

## Detailed Proof for Equation 5 in Deconfounder

**This subsection is supplemented for the detailed proof of Equation 5 in the Section "Deconfounder" in the main paper.** According to (Xu et al. 2015), we can use NWGM (i.e., normalized weighted geometric mean) to efficiently move the expectation into feature level to prevent massive calculation. We use $n_{k,i}$ to denote the logit of $k$-th candidate for $i$-th substitute confounder and $n_{k,i} = [f_n(R, \hat{g}_i, X)]_k$, where $f_n$ is the logit calculation function, as we do a softmax over all the location candidates, the original expectation can be written as:

$$\begin{aligned}
P(L|do(R), X) &= \mathbb{E}_{\hat{g} \sim \hat{G}}[P(L|R, \hat{g}, X)] \\
&\approx NWGM[P(L|R, \hat{g}, X)] \\
&= \frac{\prod_i \exp(n_{k,i})^{\hat{g}_i}}{\sum_j \prod_i \exp(n_{j,i})^{\hat{g}_i}} \\
&= \frac{\exp(\mathbb{E}_{\hat{g}}[n_{k,i}])}{\sum_j \exp(\mathbb{E}_{\hat{g}}[n_{j,i}])}, \\
&\approx P(L \mid \mathbb{E}_{\hat{g} \sim \hat{G}}[R, \hat{g}], X),
\end{aligned} \quad (8)$$

where the last approximation is because $X$ is not involved for the logit expectation over $\hat{g}$, i.e., it can be extracted from the expectation.

## Implementation Details for Deconfounder

**This subsection is supplemented for the implementation of Equation 8 in the Section "Referring Expression Deconfounder" in the main paper.** The detailed implementation of our auto-encoder in step one is shown in Table 1, and the formulation of the top-down attention is:

$$\begin{aligned}
\alpha_k &= \text{softmax}(\hat{g}_i \cdot f_r(r_k)), \\
Att(Q &= \hat{g}_i, K, V = r) = \sum_k \alpha_k r_k,
\end{aligned} \quad (9)$$

where $r_k$ is the $k$-th feature of $R$, $f_r$ is a MLP to project $r_k$ into the same dimension with $\hat{g}_i$.

## More Details for Dataset and Training Settings

**Datasets**. **This subsection is supplemented for "Datasets" in the main paper.** RefCOCOg is collected in a non-interactive setting, while RefCOCO and RefCOCO+ are collected in an interactive game interface. Three datasets have the average query lengths of 8.4, 3.6 and 3.5, respectively. RefCOCO+ avoids using absolute location words. **ReferItGame** (Kazemzadeh et al. 2014) has totally 20,000 images from the SAIAPR-12 dataset (Escalante et al. 2010), which are split by (Hu et al. 2016) into train/ validation/ test set, with 9,000/ 1,000/ and 10,000/ images, respectively. **Flickr30K Entities** (Plummer et al. 2015) has 31,783 images with 427K captions, and every caption corresponds to a certain region in the image. We use the same splits as in (Plummer et al. 2015).

**Visual Representation.** **This subsection is supplemented for "Implementation Details" in the main paper.** We followed the standard setting in (Redmon and Farhadi 2018): first resizing the long edge of images to 256 with their aspect ratio fixed; and then padding the images to the size of 256×256 using the mean pixel value. We followed the baseline one-stage method (Yang et al. 2019) to apply data augmentation and deploy the Darknet53 (YOLO-V3) (Redmon and Farhadi 2018) pre-trained on MS-COCO as our visual backbone (to extract image features). We deployed the feature-pyramid module to extract feature maps at 3 spatial resolutions: $8 \times 8$, $16 \times 16$, $32 \times 32$, and applied 3 anchors at each pixel (on the ReferItGame and Flickr30K Entities). Specially for the RefCOCO-series, we used only $8 \times 8$ feature maps (which actually produce a comparable result with that of using 3 spatial resolutions). We reduced the dimension of the visual feature to 512 by a $1 \times 1$ convolutional layer with Batch Normalization (Ioffe and Szegedy 2015) and ReLU (Glorot, Bordes, and Bengio 2011).

**Optimization.** **This subsection is supplemented for "Implementation Details" in the main paper.** We used the RMSProp (Tieleman and Hinton 2017) optimizer to learn the grounding model for 100 epochs. We set the initial learning rate as $10^{-4}$, and applied the polynomial schedule with the power as 0.9. We set the fine-tuning learning rate of the visual backbone always as $10\%$ of the above learning rate during the training process. Specially, for LBYL (Huang et al. 2021), we used Adam (Kingma and Ba 2014) optimizer and applied polynomial schedule with the power as 0.9. For MCN (Luo et al. 2020), we set the synchronous learning rate

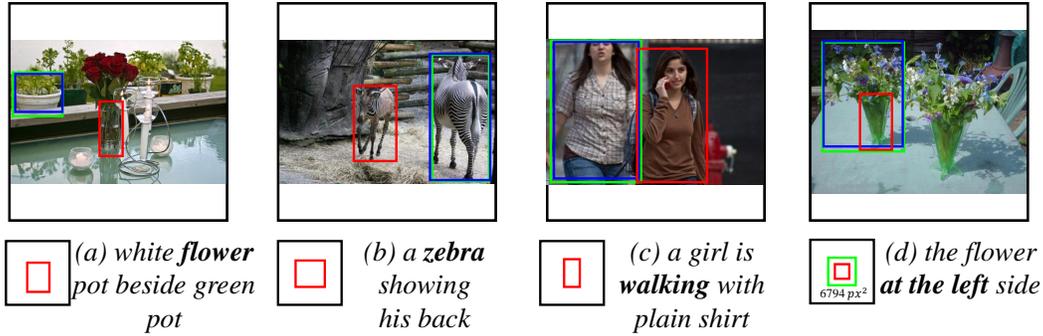

*(a) white **flower** pot beside green pot*

*(b) a **zebra** showing his back*

*(c) a girl is **walking** with plain shirt*

*(d) the flower **at the left** side*

Figure 3: More bias corrections results from RefCOCO+ (a, b, c) and RefCOCOg (d). The notations are the same as Figure 4 in the main paper (i.e., Each example contains an image, with green, blue and red boxes denoting ground truth, the predictions of Yang's-V1$^\dagger$+RED and Yang's-V1$^\dagger$, a language query at the right-bottom corner and the bias for the bold words in the query at the left-bottom corner).

| Methods | RefCOCO | | | RefCOCO+ | | | RefCOCOg | |
|---|---|---|---|---|---|---|---|---|
| | val | testA | testB | val | testA | testB | val-u | test-u |
| Yang's-V1$^\dagger$ | 74.67 | 75.98 | 70.76 | 57.21 | 61.11 | 50.93 | 61.89 | 61.56 |
| +AE($X \oplus R, X' \oplus R'$) | 78.65 | 79.42 | 75.76 | 63.46 | 66.74 | 57.23 | 69.01 | 68.55 |
| +AE($R, R'$) (RED) | **78.76** | **80.75** | **76.19** | **63.85** | **66.87** | **57.51** | **69.46** | **69.51** |
| +Att($X$)+Att($R$) | 78.30 | 79.37 | **76.36** | 63.52 | 65.57 | 56.37 | 67.83 | 68.11 |
| +Att($R$) (RED) | **78.76** | **80.75** | 76.19 | **63.85** | **66.87** | **57.51** | **69.46** | **69.51** |
| Yang's-V2 (Yang et al. 2020) | 76.59 | 78.22 | 73.25 | 63.23 | 66.64 | 55.53 | 64.87 | 64.87 |
| Yang's-V2 +RED | 78.03$_{+1.44}$ | 79.17$_{+0.95}$ | 74.91$_{+1.66}$ | 63.80$_{+0.57}$ | 66.83$_{+0.19}$ | 58.68$_{+3.15}$ | 66.87$_{+2.00}$ | 65.83$_{+0.96}$ |

Table 2: The ablation study of using different auto-encoders and different fusion methods. $X$ and $R$ denote image and language query, respectively. AE(input value, output value) denotes the auto-encoder fed with different input values for generating corresponding output values. Att(value) denotes the attention operation in using different values of $V$. The bottom two rows show the results of Yang's-V2 and the corresponding +RED.

for optimizing visual backbone (Redmon and Farhadi 2018). We conducted all experiments using Tesla V100 GPUs with the unified batch size 64.

## More Quantitative Examples

**This section is supplemented for "Quantitative Results" in the main paper.** We provide more quantitative examples in Table 2 for the ablation study on RefCOCO, RefCOCO+ and RefCOCOg datasets.

## More Qualitative Examples

**This section is supplemented for "Qualitative Results" in the main paper.** We provide more qualitative examples in Figure 3 to justify the language bias corrections of our RED.


\end{document}